\documentclass[11pt]{article}

\usepackage[final]{acl}
\usepackage{float}
\usepackage{times}
\usepackage{latexsym}
\usepackage{xfrac}
\usepackage{placeins}
\usepackage{hyperref}
\usepackage{float}
\usepackage{amsmath}

\usepackage[T1]{fontenc}
\usepackage{tcolorbox}
\tcbuselibrary{skins}
\usepackage{xcolor}
\usepackage[utf8]{inputenc}

\usepackage{microtype}
\usepackage{graphicx}
\usepackage{booktabs}

\title{Thinking effort aligns between humans and reasoning models in abductive reasoning}

\author{Henry Arthur \\
  Center for Mind/Brain Sciences, University of Trento, Italy \\
  \texttt{henryj.arthur@studenti.unitn.it}}

\begin{document}
\maketitle

\begin{abstract}
A major question in cognitive modeling concerns the behavioral alignment between large language models and humans across linguistic and non-linguistic tasks. Unlike standard LLMs, large reasoning models (LRMs) are optimized with reinforcement learning from verifiable rewards, encouraging correct solutions to reasoning tasks rather than preference-aligned responses. Recent work \cite{varda_cost_2025} investigates the cost of thinking in humans and LRMs by comparing human reaction times with model reasoning traces across a range of reasoning tasks. We isolate this alignment by turning to abductive reasoning: unlike deductive tasks, its difficulty cannot be inferred from formal structure and offers no shortcuts a model could exploit to mimic effort without genuine search, providing firmer ground for empirical claims of shared effort. We find further evidence of alignment between LRM and human reasoning effort, as well as evidence that models and humans tend to make similar errors. Finally, we show that decoding methods that let models explore multiple reasoning paths increase alignment in reasoning cost between humans and LRMs across the three models tested.
\end{abstract}

\section{Introduction}

A growing area of research investigates the plausibility of LLMs as models of human
language processing and cognition \citep{schrimpf_neural_2021, binz_foundation_2025,
piantadosi_modern_nodate, katzir_why_2023}. In general, there is disagreement about the
extent to which artificial neural network models are general solvers
\citep{lan_benchmarking_nodate, brown_language_2020}, especially for out-of-distribution
problems \citep{vargas_guzman_testing_2024}. There is an apparent tension between assessing human-model alignment across language and general reasoning, which stems partly from the assumption that linguistic competence underlies thinking.

fMRI evidence has shown a dissociation between the language network (LN) and the
multiple-demand (MD) network \citep{fedorenko_language_2024}. The LN is a distinct set of brain
regions in the left-hemisphere frontal and temporal brain areas. It is highly selective for
language and causally important for linguistic decoding and encoding. The MD network is found to
be diffusely active across a range of functions, including inductive reasoning, math, and novel
problem-solving \citep{fedorenko_language_nodate}. In a study on LLM-brain alignment over the
course of training, the LN correlated primarily with formal linguistic rules and less with world
knowledge and reasoning \citep{alkhamissi_language_2025}.

Further, empirical evidence from patients with aphasia suggests that logical reasoning does
not depend on natural language \citep{kean_evidence_2026}. Such findings may explain why LLMs
learn syntactic representations faster than world knowledge \citep{mahowald_dissociating_2024,
piantadosi_computational_2021} and struggle with logic problems \citep{malek_frontier_2025}.

Large reasoning models (LRMs) confront some of the common pitfalls in the training
regime in LLMs known as reinforcement learning from human feedback (RLHF) by adopting
a reward scheme in the fine-tuning phase called Reinforcement learning from verifiable
rewards (RLVR), which optimizes for correct answers on reasoning tasks
\citep{deepseek-ai_deepseek-r1_2025}. LRMs produce ``Chains-of-thought'', which are
linguistic traces of the reasoning process \citep{wei_chain--thought_2023}, that can be used
as a proxy for computational effort. Additionally, LRMs can
allocate more tokens during inference to solve more difficult problems
\citep{muennighoff_s1_2025, snell_scaling_2024, deepseek-ai_deepseek-r1_2025}, an
important feature of human cognition.

Although both LRMs and LLMs emit token sequences during CoT, it is still unclear if the underlying computation is linguistic. Degenerate or language-mixed traces can still confer benefits \citep{deepseek-ai_deepseek-r1_2025}, and meaningless filler tokens can substitute for CoT on some tasks \citep{pfau2024think}. Further, reasoning can proceed entirely in latent space without emitting linguistic tokens \citep{hao_training_2025}, showing the linguistic surface is separable from the computation. As noted by \citet{varda_cost_2025}, CoT outputs linguistic traces that may not be faithful to the underlying computation \citep{barez_chain--thought_nodate}, likely because
the true ’thinking’ mechanism is non-linguistic as in humans \citep{fedorenko_language_nodate}.

\subsection{Related Work}

Recently, \citet{varda_cost_2025} investigated the alignment between LRM reasoning traces
measured by token count and human reaction times across seven diverse reasoning tasks.
In testing on DeepSeek-R1, they find that models reliably track human thinking effort both
within and across tasks. They also replicate their findings to five additional open-weight
reasoning models.

This work has garnered substantial discourse in a short period of time. Some authors have
expressed skepticism about the algorithmic correspondence between the brain and LRMs
\citep{dujmovic_no_nodate, noauthor_correlations_nodate}. Importantly, the investigation of such an alignment is
not a mechanistic thesis but rather an empirical phenomenon that may account for similar
constraints across substrates. Other criticism has focused on the validity of CoT as a
measure of thinking effort \citep{hu2026thinking}, which we address in part by testing multiple decoding
strategies and reasoning effort levels (see \ref{faq:a3}).

Theoretically, such observations are relevant in the contentious debate of whether symbolic
processing can emerge from sub-symbolic architectures \citep{fodor_connectionism_1988,
smolensky_proper_nodate}.

\subsection{Abductive Reasoning}

We examine the alignment between human and model
reasoning effort in abductive inference. Introduced by \citet{peirce1965}, abduction is colloquially understood as an
educated guess or best explanation \citep{burch_charles_2024}. More formally, it is the
process of forming a hypothesis that serves as an explanation for underdetermined premises. Abduction is a natural form of reasoning that pervades daily life where people operate under uncertainty. Waking up to the sight of wet grass, one could posit several abductive explanations: my neighbor watered the grass last night, a nearby ground well began to overflow, the sprinkler system achieved sentience and rebelled, or perhaps it rained. 

It is distinct from induction and deduction and has been
argued to be a core tool for generating hypotheses about scientific explananda \citep{lipton_inference_nodate,
seddon_nature_2022} and central to reasoning between the lines in discourse \citep{hobbs_interpretation_1988}.

In human-model alignment, abduction serves as a uniquely fruitful type of reasoning to study. In syllogistic and deductive tasks, a problem's difficulty is partially recoverable from its formal structure; a model that has learned the relevant inference schemas could produce more tokens on harder-looking items without doing real search.\footnote{Thanks to Andrea de Varda for suggesting this framing (personal communication).} Abduction lacks this property: since the evidence underdetermines the conclusion by design, no such shortcut is available, providing stronger evidence of shared effort. As an example, consider two syllogisms of similar surface length:

\begin{itemize}
    \item[(1)] All A are B; all B are C; therefore all A are C. (easy mood)
    \item[(2)] Some A are not B; all C are B; therefore some A are not C. (harder mood)
\end{itemize}

A model that has learned which forms are hard could allocate more tokens to (2) reflecting memorized structural difficulty cues rather than shared effort. In our abductive task, there are no formal markers that would suggest which answer is correct or how difficult the discrimination is. Difficulty arises only from searching over what each hypothesis would explain.

\subsection{Abduction in LLMs}

The study of LLMs' abductive
ability has recently attracted interest (for a review, see \citealp{salimi_wiring_2026}). Studies have
covered abduction in syllogistic form \citep{abe_abductive_2026}, commonsense hypothesis selection/generation \citep{bhagavatula_abductive_2020}, and medical diagnosis \citep{wu_medcasereasoning_2025}. Most previous work has observed that LLMs struggle with abduction and are
consistently outperformed by humans. Explicit inference to the best explanation criteria has also been shown to improve model-based evaluation of commonsense explanations \citep{dalal-etal-2024-inference}.

\subsection{Abductive Reasoning Dataset}
We operationalize abduction in a forced-choice task in which two candidate hypotheses are
presented as possible explanations for two observations, in the style of inference to the best
explanation. In an online experiment, we collected data from 120 participants, recording
accuracy and reaction time. The experimental items were drawn from \citet{bhagavatula_abductive_2020},
who created a commonsense dataset adapted from ROCStories \citep{mostafazadeh_corpus_2016},
a corpus of narrative short stories. \citet{bhagavatula_abductive_2020} used the beginnings and
endings of stories as observations and recruited human participants to generate hypotheses
explaining the change of state from O1 to O2. They had a different set of participants revise
the hypothesis in a minimal way that would make one hypothesis wrong. The final step was to use BERT
as an adversarial filter to select hypothesis pairs that were most difficult to discern as the
correct answer.

\subsection{Decoding}

The decoding methods in LLMs have been widely observed to affect reasoning patterns and
accuracy \citep{wang_self-consistency_2023, wang_chain--thought_2024,
wang_contextual_nodate}. See \citet{renze_effect_2024} for opposing evidence. Temperature
is a hyperparameter in language models that adjusts the target distribution produced by the
final softmax layer; a higher temperature yields a softer distribution over candidate tokens
\citep{hinton_distilling_2015}. Decoding strategies that optimize for high-probability outputs
can lead to degeneration \citep{holtzman_curious_2019}. Thus, running reasoning models at lower temperatures can increase the risk of thinking loops, in which models produce repetitive or near-identical reasoning traces, whereas higher temperatures can avoid looping
\citep{pipis_wait_2025}. Model providers suggest running models at sufficiently high
temperatures to encourage exploration of alternative reasoning paths and to avoid loops;
these temperatures usually range from 0.6 to 1 \citep{deepseek-ai_deepseek-r1_2025,
yang_qwen3_2025, openai_gpt-oss-120b_2025}.

The paper is structured as follows. We first introduce the dataset and experimental setup,
including our adaptation of \citeauthor{varda_cost_2025}'s method to abductive reasoning under an
inference-to-the-best-explanation paradigm, before presenting the main results. We then
examine non-greedy decoding, showing that temperature-based sampling aggregated
across multiple runs strengthens human--model correlations. Finally, we situate these
findings within previous work on thinking-cost alignment and discuss why the decoding
strategy may matter for measuring reasoning effort.

Since human reaction times are strongly influenced by item length, all reported token–RT \textit{r}s are partial
correlations controlling for problem length.

\section{Methods}

We tested participants ($N = 120$) across 160 items from the Abductive Natural Language
Inference Dataset from AI2 in an online forced-choice experiment. Participants were
instructed to read two observations that described commonsense scenarios and select one
of two hypotheses that best explained both observations. Participants were instructed to
answer as quickly as they could while remaining accurate.

\begin{figure}[h]
\centering
\begin{tcolorbox}[
    colback=gray!6,
    colframe=black!70,
    width=\linewidth,
    arc=2pt,
    boxrule=0.6pt,
    top=4pt, bottom=4pt, left=6pt, right=6pt
]
\scriptsize
\textbf{Observation 1:} Alec works at the renaissance fair.\\
\textbf{Observation 2:} They had to make it seem like part of the act.\\[3pt]
\textbf{Hypothesis A:} Alec jumped up in front of everyone.\\
\textbf{Hypothesis B:} Alec's phone started ringing during a performance.\\[3pt]
\textbf{Correct answer: Hypothesis B}
\end{tcolorbox}
\caption{Example item from the abductive forced-choice task.}
\label{fig:example}
\end{figure}

\subsection{Participants}

Participants ($N = 127$) were recruited through Prolific. All participants were
Native English speakers residing in the United States with normal to corrected-to-normal
vision. A total of 7 participants were removed for implausibly fast responses or
disproportionately long responses, indicating they had left their computer. The final sample
includes 120 participants (mean age = 33.0 years, SD = 6.6; 60\% Female, 38\% Male, 2\%
Other).

\subsection{Design}

Participants were divided into 4 blocks; each block contained 40 unique
abductive scenarios, covering 160 questions. In the analysis, Block 1 ($N=28$), Block 2
($N=31$), Block 3 ($N=32$), Block 4 ($N=29$), with no overlapping problems between blocks.

\subsection{Procedure}

The survey was conducted in English. Participants used the keyboard to
choose between hypotheses on-screen using the Q and P keys. Before beginning the task,
participants were tested with two multiple-choice questions to ensure they understood the
instructions. Two practice problems were provided to help participants familiarize themselves
with the layout, and they were instructed to keep their fingers on P and Q throughout the
experiment. The answer to each question was randomized between P and Q to avoid
biasing participants towards one key. Two attention trials were added to catch participants
who were unfocused. A fixation cross appeared briefly between trials (200ms), and time was
unlimited for each problem. The mean completion time for the study was 9.2 minutes (SD =
4.0).

\subsection{Materials}

The ART dataset from AllenAI consists of a multiple-choice
question-answering task. Each item
consists of two observations (O1, O2) and two hypotheses (A, B), with the task being to
select the hypothesis that better explains the observations. The data is publicly available on \href{https://huggingface.co/datasets/allenai/art}{HuggingFace} and consists of training and test sets. To prevent participants from using
response-option length as a heuristic, we retained only items for which hypotheses A and B
differed by at most 1 word. Duplicate contexts (identical O1--O2 pairs) were removed. From
the resulting pool, 162 items were randomly sampled (seed = 42): 160 main items and 2
practice items. All items came from the test set, where items underwent an adversarial
filtering algorithm to retain hypotheses that are harder to distinguish, which demands deeper
reasoning from participants and models. Practice items were selected as the two shortest
items (by total word count) in the sample to serve as clear warm-up examples and were the
same across all blocks. The 160 main items had a mean total word count of 33.5 (SD = 7.0,
range = 19--70) across both observations and both hypotheses. Of the 160 items, 68 had
hypotheses of equal length, and 92 differed by exactly one word. The correct answer was
option A on 87 items and option B on 73 items.
\begin{figure}[h]
\centering
\begin{tcolorbox}[
    colback=gray!6,
    colframe=black!70,
    width=\linewidth,
    arc=2pt,
    boxrule=0.6pt,
    top=6pt, bottom=4pt, left=6pt, right=6pt,
    attach boxed title to top left={xshift=6pt, yshift=-4pt},
    boxed title style={
        colback=black!70,
        colframe=black!70,
        arc=1pt,
        boxrule=0.4pt,
        top=3pt, bottom=3pt, left=8pt, right=8pt
    },
    title={\textcolor{white}{\bfseries\scriptsize Prompt}}
]
\scriptsize
In this task, you will read two observations about a situation, followed by two possible explanations (hypotheses). The observations describe the beginning and end of a short story. Your task is to decide which hypothesis more plausibly explains what happened in between. Read the observations and hypotheses carefully, then select the hypothesis that best connects Observation 1 to Observation 2. Your primary goal should be to give the correct answer, but try to do so as quickly as possible.\\[3pt]
\textbf{Observation 1:} \textit{[O1]}\\
\textbf{Observation 2:} \textit{[O2]}\\[3pt]
\textbf{Hypothesis A:} \textit{[A]}\\
\textbf{Hypothesis B:} \textit{[B]}\\[3pt]
Which hypothesis (A or B) more plausibly explains what happened between Observation 1 and Observation 2?\\[2pt]
Format your answer like this: \texttt{`Answer: A'} or \texttt{`Answer: B'.}\\[2pt]
\textcolor{gray}{\textit{[V3 only]:} Let's think step by step.}
\end{tcolorbox}
\caption{Prompt template used across conditions. Placeholders in italics are filled with each item's content. The grey line was appended only in condition V3.}
\label{fig:prompt}
\end{figure}
\subsection{Model evaluation}
Following the method of \citet{varda_cost_2025}, we analyze seven large reasoning models (LRMs) and one non-reasoning baseline: DeepSeek R1 (hereafter R1), Qwen3-235B (Thinking), Qwen3-32B (Thinking), GPT-OSS-20B, GPT-OSS-120B, GLM-4.5-Air, Kimi-K2 (Thinking), and DeepSeek V3. Five of these were included in the original analysis; we substituted Qwen3-32B for QwQ-32B, which was unavailable, and added Kimi-K2 to test robustness across reasoning models of differing size. We use the OpenRouter API with greedy decoding (temperature = 0) for all models and default reasoning effort (see Appendix \ref{appendix:hyperparameters}). The prompt was designed to mirror instructions given to human participants. For LRMs, responses include a thinking trace enclosed in \texttt{<think>...</think>} tags, which we use to count tokens generated during reasoning; for DeepSeek V3, ``let's think step by step'' \citep{kojima_large_2023} is appended to the prompt to elicit chain-of-thought reasoning, since it is not native to the model. Reasoning effort is measured as the total tokens generated for DeepSeek V3.

We also expand the previous work by testing GPT-OSS-20B, R1, and Qwen3-32B at temp
0.6 and temp 1 as suggested by model-provider guidelines
\citep{deepseek-ai_deepseek-r1_2025, yang_qwen3_2025, openai_gpt-oss-120b_2025} to
elicit better reasoning patterns. We run GPT-OSS-20B and Qwen3-32B 25 times each, and
10 times for R1, and aggregate over item-level responses to obtain the majority-vote answer label and mean token cost. To determine the minimum number of independent runs
required for a stable partial correlation estimate, we ran a convergence bootstrap analysis.
For each value of $k$ (number of runs averaged, from 1 to $K-1$), we drew 500 random
subsets of $k$ runs, computed the partial correlation between mean reasoning tokens and
human RT (controlling for prompt length) for each subset, and recorded the mean and 95\%
CI across subsamples. Convergence was declared at the first $k$ for which the 95\% CI
width fell below 0.03, and the absolute change in mean partial $r$ from $k-1$ to $k$ fell
below 0.01. To see whether higher temperatures adversely affect
reasoning, we ran the three models at temp = 2,
using the same number of runs. Because bootstrap CI shrinks mechanically as k approaches the full pool, we cross-validated our convergence estimates by exploiting the fact that chain-of-thought length converges as $1/\sqrt{k}$, solving analytically for the number of runs needed to bring the standard error of log-transformed token counts below a fixed threshold (Appendix \ref{appendix:cot-convergence}).


\section{Results}

We analyzed the variance in the forced-choice test to determine whether there is a
correlation between thinking models and human reasoning effort. We computed Pearson
correlations between the number of reasoning tokens per model and the average human
RTs, both log-transformed and per-item. Since item length is highly correlated with reaction
time in our study ($r = 0.669$, $p < .001$), we use partial correlations to isolate thinking time
(residualizing log RT and log reasoning tokens on log prompt token count, as returned by
each model's native API tokenizer). Regression lines show ordinary least-squares fits. Our
results reflect the correlation between incorrect and correct items combined for models and
humans. An analysis of all responses (model \& human), human-correct RT only, and human-correct and model-correct only did not provide significantly different
correlations for any model. Results were robust to the operationalization of prompt length:
rerunning partial correlations using log word count instead of log token count yielded no
significant differences for all models except GPT-OSS-120B, for which the token-based
partial correlation was marginally higher ($\Delta = +0.08$, 95\% CI $[+0.005, +0.161]$). In
both cases, GPT-OSS-120B remained highly significant ($p < .001$), and the direction and
ordering of results were unchanged.
\begin{figure}[t]
  \centering
    \includegraphics[width=0.8\columnwidth]{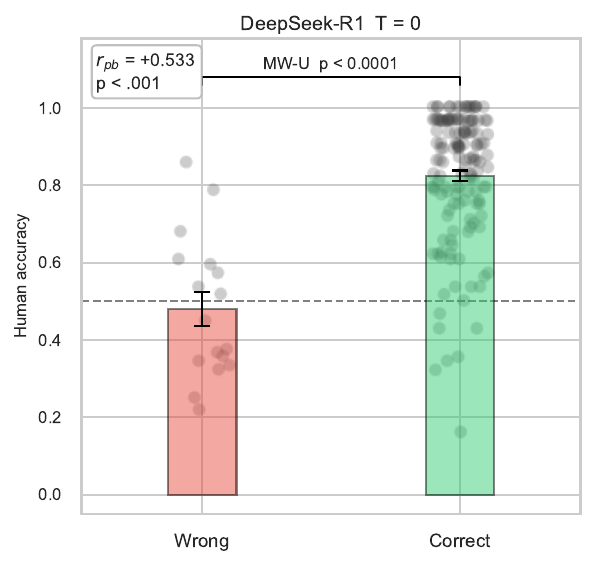}
  \caption{Human accuracy on items DeepSeek-R1 answered correctly versus incorrectly (T=0, greedy). Items the model got wrong corresponded to substantially lower human accuracy.}
  \label{fig:errorsim}
\end{figure}

\begin{figure*}[tbp]
  \centering
    \includegraphics[width=0.9\linewidth]{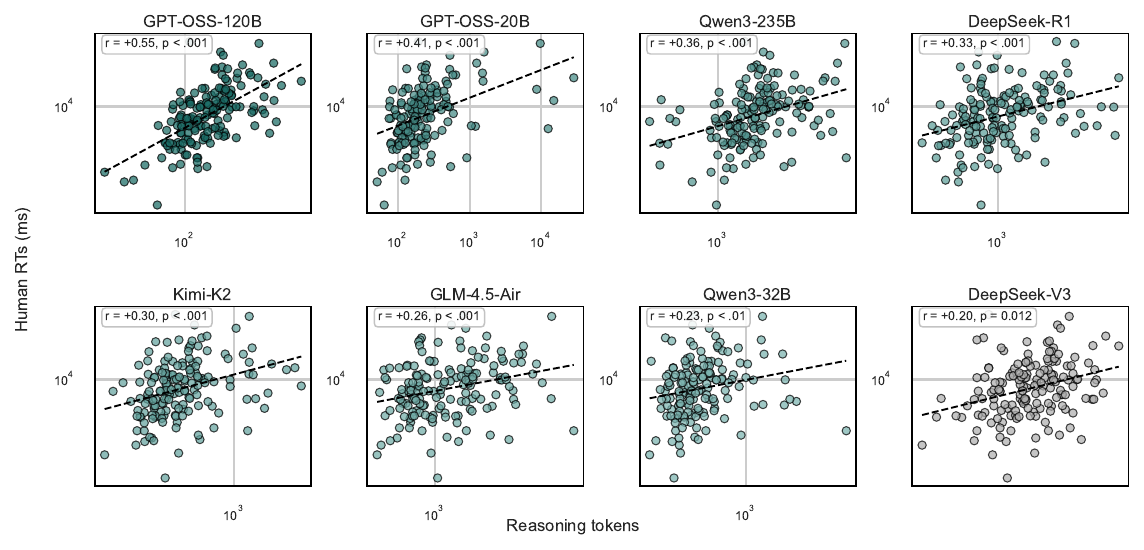}
  \caption{Reasoning tokens versus mean human RT for all models tested under Greedy decoding; each dot represents one abductive item.}
  \label{fig:example}
\end{figure*}

\subsection{Greedy Decoding Correlations}

After controlling for prompt length, all models are significantly correlated with human
reasoning effort. The strongest alignment was observed for GPT-OSS-120B ($r = 0.552$, $p
< .001$) and GPT-OSS-20B ($r = 0.41$, $p < .001$), followed by Qwen3-235B-Thinking
($r = 0.359$, $p < .001$), DeepSeek-R1 ($r = 0.327$, $p < .001$), Kimi-K2-Thinking ($r =
0.303$, $p < .001$), GLM-4.5-Air ($r = 0.263$, $p < .001$), and Qwen3-32B-Thinking ($r =
0.229$, $p = .002$). The only non-reasoning model, DeepSeek-V3, was also significantly
correlated ($r = 0.197$, $p = .012$), though it showed the weakest alignment of all models
tested. Notably, the difference between DeepSeek-R1 and its non-reasoning counterpart
DeepSeek-V3 did not reach significance (Fisher $z = 1.23$, $p = .217$), suggesting that
while reasoning models tend to show numerically stronger alignment, the advantage over a
capable non-reasoning model was not reliably detected at this sample size. As a robustness
check, we recomputed all partial correlations using Spearman rank correlation (see Appendix \ref{appendix:correlations}). Spearman partial $\rho$ values closely matched Pearson partial $r$ values across all
models and conditions, confirming that the relationships are monotonic and not driven by
distributional assumptions of linear correlation.

\subsection{Greedy Error Alignment}

We were also interested in whether models and humans succeeded on similar items. Model
consensus accuracy was the simple unweighted proportion of models answering each item
correctly, with all models contributing equally. Mean human accuracy was 78.8\%. Looking at
DeepSeek-R1, human accuracy on items R1 answered correctly was 82.5\%, compared to
47.8\% -- below chance -- on items R1 got wrong (point-biserial $r = 0.53$, $p < .001$; see Figure~\ref{fig:errorsim}).
This pattern generalized across all models: the correlation between model consensus
accuracy and human accuracy, computed as the Pearson partial correlation across the 160 items between per-item model-consensus accuracy and per-item human accuracy, controlling for prompt length, was $r = 0.668$ ($p < .001$). When at least 75\% of models
failed an item, human accuracy was 48.7\%; when at least 75\% of models agreed on the
correct answer, human accuracy was 84.6\%. This shared sensitivity to item difficulty argues
against a simple memorization account: if these items were present in model pretraining,
near-ceiling performance would be expected. The presence of systematic, human-aligned
errors instead suggests the models are reasoning about the problems.

\begin{figure*}[t]
  \centering
    \includegraphics[width=0.9\linewidth]{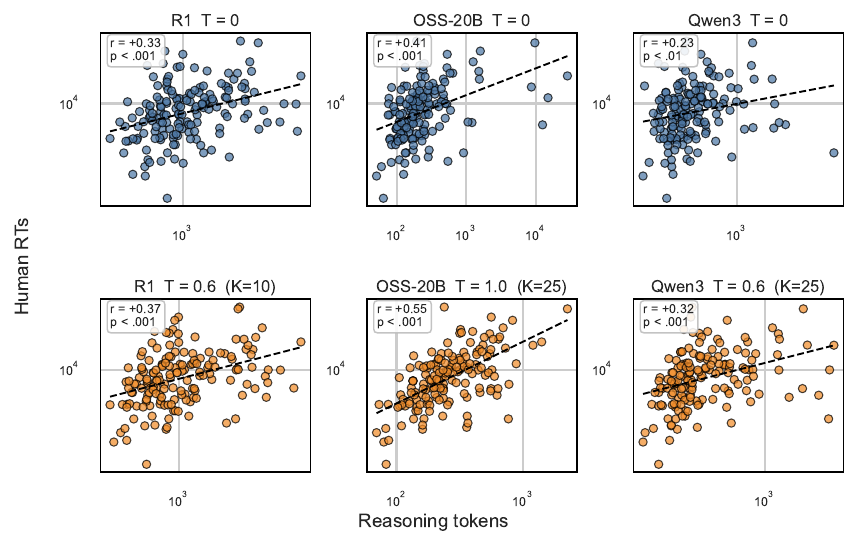}
  \caption{DeepSeek-R1, GPT-OSS-20B, and Qwen3-32B under greedy decoding (T=0, top row) and stochastic decoding (averaged across K runs, bottom row). Alignment increases reliably under stochastic sampling.}
  \label{fig:temp}
\end{figure*}

\subsection{Stochastic Decoding Correlations}

\begin{figure}[t]
  \centering
    \includegraphics[width=\columnwidth]{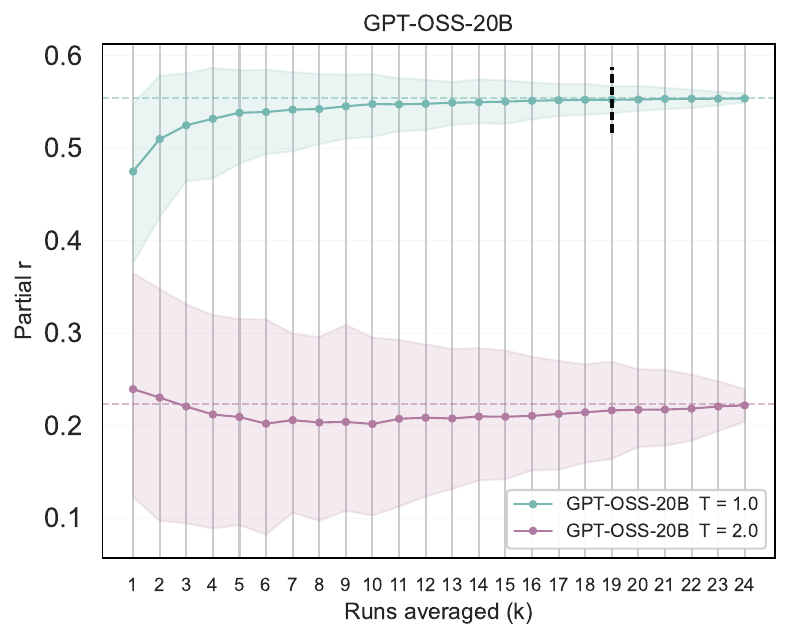}
  \caption{Convergence of the partial-r estimate as a function of the number of stochastic runs (K) for GPT-OSS-20B at T=1.0 and T=2.0. Shaded bands indicate 95\% bootstrap confidence intervals.}
  \label{fig:converge}
\end{figure}

To analyze the effects of decoding on reasoning alignment, we ran three models with non-greedy decoding. First, we used the temperatures recommended by the model providers:
R1 (temp = 0.6), GPT-OSS-20B (temp = 1), and Qwen3-32B (temp = 0.6). For each model, we
ran it many times and computed the mean token response and answer label.\footnote{The number of stochastic runs was determined iteratively: we chose an initial K a priori and increased it until estimates converged (final values: K = 10 for R1, K = 25 for GPT-OSS-20B and Qwen3-32B).} We assess
stability of the pooled partial correlation across runs by Monte Carlo subsampling: for each
$k$, we repeatedly draw $k$ runs at random, recompute partial $r$, and summarize the
bootstrap distribution's width; we declare convergence at the smallest $k$ (strictly below
the full pool) where the 95\% interval is sufficiently narrow and the mean partial $r$ changes
little when adding another run. Using the
recommended temperatures and aggregating model responses improved partial
correlations across all models. In GPT-OSS-20B, the partial went from $r =0.41$ to $r=0.55$, (Steiger–Williams test,
$p < .001$), and the model converged to a stable partial at $k=19$ runs. For Qwen3-32B,
greedy $r = 0.23$ and at temp = 0.6, $r=0.32$ $(p < 0.05)$,
and converged at 18 runs. R1 also improved from greedy decoding at $r = 0.32$ and $r =
0.37$ for temp = 0.6, but the difference was not significant, and this model converged after
8 runs. Across all three models, stochastic averaging consistently improved alignment with human RT, significantly so for GPT-OSS-20B (p < .001) and Qwen3-32B (p < .05). We were also curious how the models would align at high, non-recommended
temperatures, so we tested all three models again at temp = 2 for the same number of
runs. GPT-OSS-20B partial $r$ decreased to 0.22 ($\Delta$= $-$.19), and the partial did not
converge after 25 runs (Figure ~\ref{fig:converge}). Qwen3-32B partial increased by $\Delta$ = $+$.063, but was still lower
than the results with temp = 0.6 and converged at 24 runs, 6 runs more than temp 0.6. R1
Temp = 2 increased from T=0 by $\Delta$ = $+$.028 and converged after 6 runs.\footnote{Token
variance across stochastic runs did not predict human RT beyond what mean token count
already explained (partial $r$s = $-.12$ to $+.08$, all $p$s $> .12$), confirming that a
simple per-item mean is sufficient and that results are not driven by a small number of
high-variance outlier runs.}

\subsection{Stochastic Error Alignment}

We also analyzed the alignment with human accuracy across 
recommended temperature settings and high temperatures. We use a point-biserial
correlation between model responses (correct/wrong) and human accuracy \% across all
participants and items. For the model GPT-OSS-20B, temp=1 had the highest correlation with
human accuracy/errors ($r = 0.51$), followed by $r = 0.5$ at temp = 2 and $r = 0.42$ at
temp = 0, all $p$s $< .001$. For R1, the correlation was stronger at temp = 0 ($r = 0.53$)
than at temp = 0.6 ($r = 0.51$), and lowest for T = 2 at $r= 0.48$, all $p$s $< .001$.
Qwen3 showed the strongest correlation for t = 0.6 with $r=0.63$, temp = 0 with $r=0.6$, and at
temp = 2 with $r = 0.54$, all $p$s $< .001$. Even though Qwen and R1 had a higher partial at
temp = 2 than temp = 0, their correlation with human accuracy was the lowest.

\section{Discussion}

Our results fall within the range expected from \citet{varda_cost_2025}. Averaged across models under greedy decoding, the ensemble partial correlation (mean log tokens per item) was $r = 0.42$, $p < .001$, which closely matches their within-task result for syllogisms ($r = 0.43$) and exceeds relational reasoning ($r = 0.27$). This comparison is theoretically meaningful because abductive forced-choice
reasoning, syllogisms, and relational reasoning all involve inference over structured, grounded scenarios. However, abduction differs from deduction in that the evidence is
underdetermined: the participant must select the best explanation rather than derive a
conclusion that follows from sufficient premises. For this reason, it is unsurprising that
abductive reasoning falls near other scenario-based reasoning tasks, rather than arithmetic
or formal logic.

Importantly, all LRMs outperform V3, the non-reasoning LLM tested here, suggesting that
chain-of-thought traces in LRMs capture thinking cost in a way that is not reducible to
ordinary next-token generation. Human-like sensitivity to content effects in LLMs may
further explain why models and humans show similar item-level difficulty patterns in such
tasks \citep{lampinen_language_2024}.

To assess how close model alignment comes to the maximum possible given measurement noise, we estimated a noise ceiling using bootstrapped split-half reliability of human RTs (Spearman-Brown corrected,
$ r = .861$). The strongest model, GPT-OSS-120B, explains $R^2 = .303$ of the residualized item-level RT variance, or 35\% of the noise-ceiling-explainable variance. This constitutes a substantial degree of shared variance between two fundamentally different systems, which merits further investigation.

We demonstrate that running models at their recommended temperatures for reasoning
increases the alignment between human and model reasoning cost in forced-choice
abduction.
These results suggest that stochastic decoding can be used reproducibly if convergence is
quantified across runs. Higher temperatures introduce more variance, so more runs are
required before the item-level correlation stabilizes. Averaging across runs is therefore analogous to pooling human reaction times across 
participants: it reduces noise from idiosyncratic samples (see Figure~\ref{fig:stochastic-attenuation}). 
This may be especially important for reasoning models, where generation-time errors 
can compound over the reasoning trace \citep{pipis_wait_2025}, making unstable runs 
analogous to distracted or unusually slow participants.

We also test temperatures that are typically considered too high for reasoning. In particular,
temp = 2 is rarely recommended outside of more creative generation settings. We find that
at temp = 2, model correlations are less stable, take longer to converge, and remain lower
than those obtained with recommended settings such as temp = 0.6. Across both
model-human accuracy alignment and partial correlations with human reaction times, the
recommended temperature produces the strongest alignment.

Notably, model size did not predict alignment: GPT-OSS-20B (20B parameters) showed the
second-strongest correlation despite being over an order of magnitude smaller than DeepSeek-R1
(671B parameters) or Kimi-K2 (1T parameters), both of which showed weaker correlations.

An important question arises from our results: why does stochastic decoding yield better
alignment with the cost of human reasoning?

Greedy decoding produced reliably more token outliers than stochastic sampling, with some
items triggering extended thinking loops and very high token counts. However, because
tokens were log-transformed, these extreme values had limited leverage on the correlations;
removing the worst outliers did not meaningfully change the partial $r$. Greedy sampling also showed
a higher raw token SD across all three models (DeepSeek-R1: 759 vs 613; GPT-OSS-20B:
2801 vs 224; Qwen3-32B: 558 vs 342), indicating greater overall spread. Together, these
results suggest that outliers alone do not account for greedy's weaker alignment; the log
transform largely neutralizes their influence. Instead, the stochastic advantage appears to
reflect attenuation. Single-run greedy estimates contain more noise per item, and this
noise attenuates the correlation with human RT across the full difficulty range rather than at
any particular point.

\section{Conclusion}

We investigated the alignment between large reasoning models and human thinking effort in
a commonsense abductive reasoning task, where difficulty cannot be inferred from formal structure alone. Across all reasoning models, there was a
significant correlation, and alignment was stronger with LRMs than with the LLM
DeepSeek-V3. Additionally, models and humans succeeded and failed on similar abductive
items.

In conjunction with results from other reasoning domains, we show that chain-of-thought
length can serve as a proxy for cognitive effort. We make the novel contribution of
applying non-greedy decoding methods to show that stochastic decoding, when averaged
over many runs, reveals stronger alignment between human and model thinking effort. This
suggests that within-task alignment in \citet{varda_cost_2025} may be stronger than
originally reported.

While the weights of off-the-shelf models are fixed, hyperparameters can be treated as
experimental variables to be tuned to avoid adverse effects -- analogous to how we control
environmental conditions in human experiments. Alternatively, stochastic decoding may reflect a cognitive process that is more similar in humans than greedy decoding.

\section{Future work}

Future work could run stochastic sampling across a wider range of reasoning models to
better understand what drives the differential increase in alignment across models.
Hyperparameters such as top\_p and top\_k were kept at default values here, but future
work could examine their effects.

It would also be useful to adapt our method to other abductive reasoning datasets, such
as long-form detective settings, which require deeper reasoning and the ability to
connect clues across context \citep{del_true_2023}.

There is also potential to measure alignment with CoT traces beyond linguistic output,
such as continuous latent space \citep{hao_training_2025}, or with mechanistic
properties of reasoning effort, such as token-revision counts in deep layers
\citep{chen_think_2026}.

\section*{Limitations}

We found that effort increased correctness on the hardest tercile of
problems (see Appendix~\ref{faq:a3}). All main analyses used each
model's default reasoning-effort setting rather than tuning it to
difficulty — a bias set a priori, much as a human told in advance to
expect an easy, medium, or hard problem would shift their response
speed. This suggests alignment may be stronger with datasets
calibrated to avoid ceiling effects, echoing
\citet{lampinen-2024-language}'s point that fair LLM-human comparison
requires matching the conditions faced by human participants.

We tested one type of abductive reasoning, in the style of inference to the best explanation.
However, this does not represent all forms of abductive reasoning, and important differences
in hypothesis selection and generation are not captured here.

In our model runs, we used a prompt as close as possible to human directions; however,
LLM outputs are known to be sensitive to prompts \citep{zhao_calibrate_2021,
sclar_quantifying_2024}, and variation in instructions could affect our results.

Although shared error patterns with humans suggest that ground-truth labels were not simply memorized, we cannot rule out the possibility that models encountered ART items during pretraining, which may inflate accuracy through familiarity rather than genuine reasoning.

\section*{Ethical Considerations}

The observed alignment does not imply equivalence between models and humans. Our focus is evidence of shared computational constraints, which matters for cognitive modeling but does not license anthropomorphizing or increased trust in model outputs.

No personally identifiable information was collected as part of the dataset. All participants
were recruited and compensated through Prolific and gave informed consent for their
anonymized behavioral data, including response accuracy and reaction times, to be used in
a study of human reasoning.


\section*{Acknowledgments}

Thanks to Andrea de Varda, Jakub Szymanik, and the anonymous reviewers for helpful discussions.

\section*{Data Availability}

{\small\sloppy Code and data can be found at:\\
\url{https://anonymous.4open.science/r/CostAbductiveReasoning-7052/README.md}}

\bibliography{Final_References}

\appendix
\section{Frequently asked questions}
\label{appendix:faq}
\subsection{The difference between DeepSeek-R1 and DeepSeek-V3 is not statistically significant. Does this undermine the claim that LRMs show stronger alignment with human reasoning effort?}
\label{faq:a1}
Not necessarily. Commonsense abductive reasoning relies heavily on plausible, linguistically-grounded scenarios well-represented in pretraining corpora, meaning a capable LLM like V3 can approximate human performance through world knowledge alone without explicit reasoning chains. Notably, V3 still showed the weakest alignment of all models tested. The gap between LRMs and standard LLMs is likely to become more pronounced on harder abductive problems that rely less on surface linguistic patterns, such as long-form detective reasoning, where extended chain-of-thought would provide a clearer advantage over exclusive next-token prediction.

\subsection{Why does alignment not scale with model size?}
\label{faq:a2}
All models tested (with the exception of Qwen3-32B) use a Mixture-of-Experts architecture \cite{shazeer2017outrageouslylargeneural}, where only a fraction of parameters are activated per token. Across models, activated parameter counts are actually fairly similar, so raw size comparisons may not impact this task meaningfully. Differences in alignment more likely reflect training dynamics particular to each provider, such as differences in RLVR reward schemes and fine-tuning curricula. It is also worth noting that commonsense reasoning tasks may not be the best setting to detect scale effects, since all models can draw on world knowledge from pretraining. Tasks that rely less on linguistic and commonsense patterns may reveal a clearer relationship between model characteristics and human alignment.

\subsection{Do the results change under different reasoning-effort parameters?}
\label{faq:a3}
We tested GPT-OSS-20B at low and high reasoning effort across all 160 items. Token--RT alignment held at both settings (low $r = +.34$, high $r = +.30$, both $p < .001$; Steiger's test $p = .60$). Aggregate accuracy was similar (82.5\% vs.\ 86.3\%). Stratifying by human difficulty, high effort outperformed low effort on the hardest tercile ($n = 52$: 69.2\% vs.\ 53.8\%; McNemar exact $p = .022$) but not the easiest, with the advantage growing with item difficulty (effort $\times$ human-accuracy interaction, item-clustered SEs: $\beta = -3.71$, $p = .016$). These results further suggest that reasoning alignment is clearer on hard problems and that hyperparameters should be chosen faithfully a priori.

\section{Model Specifications}

\paragraph{DeepSeek-R1} is built on DeepSeek-V3 and uses reinforcement learning to optimize for correct answers relative to ground-truth labels on verifiable tasks such as math \citep{deepseek-ai_deepseek-r1_2025}.

\paragraph{DeepSeek-V3} is a Mixture-of-Experts model with 671B total parameters, of which 37B are activated \citep{deepseek-ai2024deepseekv3}.

\paragraph{GPT-OSS-20B / 120B} are both Mixture-of-Experts models; the larger model has 128 experts, and the smaller has 32 \citep{openai_gpt-oss-120b_2025}.

\paragraph{Qwen3-235B / 32B} Qwen3-235B is a Mixture-of-Experts model, and Qwen3-32B uses a dense causal architecture with a dedicated thinking mode that can be turned on/off \citep{yang_qwen3_2025}.

\paragraph{GLM 4.5 Air} is a Mixture-of-Experts model with 355 billion total parameters and 32 billion activated parameters \citep{team_glm-45_2025}.

\paragraph{Kimi-K2} is a Mixture-of-Experts model with 1 trillion total parameters and 32 billion activated parameters \citep{team_kimi_2026}.

\section{Hyperparameters}
\label{appendix:hyperparameters}
\begin{table}[H]
\centering
\footnotesize
\resizebox{\columnwidth}{!}{%
\begin{tabular}{lccccc}
\toprule
\textbf{Model} & \textbf{Greedy} & \textbf{Stoch.} & \textbf{Top P} & \textbf{Top K} & \textbf{Effort} \\
\midrule
DeepSeek-V3  & 0 & n/a & \textit{1.0}  & n.s.p. & n/a \\
DeepSeek-R1  & 0 & 0.6 & \textit{0.95} & n.s.p. & n/a \\
GPT-OSS-120B & 0 & n/a & \textit{1.0}  & n.s.p. & medium \\
GPT-OSS-20B  & 0 & 1   & \textit{1.0}  & n.s.p. & medium \\
GLM 4.5      & 0 & n/a & \textit{0.95} & \textit{40} & n/a \\
Kimi-K2      & 0 & n/a & n.s.          & n.s.p. & n/a \\
Qwen3-235B   & 0 & n/a & \textit{0.95} & \textit{20} & default \\
Qwen3-32B    & 0 & 0.6 & \textit{0.95} & \textit{20} & default \\
\bottomrule
\end{tabular}%
}
\caption{Hyperparameters per model. Italics = provider default, not set in code. n.s.\ = not specified in code; n.s.p.\ = not specified in code or by provider; n/a = not applicable.}
\label{tab:hyperparams}
\end{table}

\section{Pearson and Spearman Correlations}
\label{appendix:correlations}
\begin{table}[H]
\centering
\resizebox{\columnwidth}{!}{%
\begin{tabular}{llccccc}
\toprule
\textbf{Model} & \textbf{Condition} & \textbf{K} & \textbf{Partial \textit{r}} & \textbf{\textit{p}} & \textbf{Spearman $\rho$} & \textbf{\textit{p}} \\
\midrule
DeepSeek-R1  & Greedy T=0  & 1  & +0.327 & $< .001$ & +0.301 & $< .001$ \\
DeepSeek-R1  & Stochastic  & 10 & +0.384 & $< .001$ & +0.356 & $< .001$ \\
GPT-OSS-20B  & Greedy T=0  & 1  & +0.41 & $< .001$ & +0.373 & $< .001$ \\
GPT-OSS-20B  & Stochastic  & 25 & +0.539 & $< .001$ & +0.526 & $< .001$ \\
Qwen3-32B    & Greedy T=0  & 1  & +0.230 & $< .002$ & +0.324 & $< .001$ \\
Qwen3-32B    & Stochastic  & 25 & +0.327 & $< .001$ & +0.376 & $< .001$ \\
\bottomrule
\end{tabular}}
\caption{Partial correlations (Pearson and Spearman) for models under greedy and stochastic decoding conditions.}
\label{tab:correlations}
\end{table}

\section{Human vs Model Accuracy}
\label{appendix:accuracy}
\begin{table}[H]
\centering
\small
\begin{tabular}{lcc}
\toprule
\textbf{Model} & \textbf{N} & \textbf{Accuracy (95\% CI)} \\
\midrule
DeepSeek-R1          & 160 & 89.4\% {[}83.6\%, 93.3\%{]} \\
Kimi-K2-Thinking     & 160 & 88.1\% {[}82.2\%, 92.3\%{]} \\
GLM-4.5-Air          & 160 & 86.9\% {[}80.8\%, 91.3\%{]} \\
GPT-OSS-20B          & 160 & 85.0\% {[}78.7\%, 89.7\%{]} \\
Qwen3-235B-Thinking  & 160 & 84.4\% {[}78.0\%, 89.2\%{]} \\
Qwen3-32B-Thinking   & 160 & 84.4\% {[}78.0\%, 89.2\%{]} \\
DeepSeek-V3          & 160 & 83.8\% {[}77.3\%, 88.7\%{]} \\
GPT-OSS-120B         & 160 & 82.5\% {[}75.9\%, 87.6\%{]} \\
\midrule
Human                & 160 & 78.8\% {[}75.6\%, 81.9\%{]} \\
\bottomrule
\end{tabular}
\caption{\small Accuracy on the abductive reasoning task under greedy decoding
    ($T=0$), sorted by model. Model CIs are Wilson score 95\% intervals
    ($n=160$); human CI is a bootstrap 95\% interval (2{,}000 resamples).}
\label{tab:accuracy}
\end{table}

\section{Chain-of-Thought Convergence}
\label{appendix:cot-convergence}

We estimated per-item token variability $\bar{\sigma}$ from our existing runs and solved analytically for the number of runs $k^{*}$ needed to bring the standard error below a threshold $\theta$ in log-token space ($\text{SE} = \bar{\sigma}/\sqrt{k}$), i.e. $k^{*} = (\bar{\sigma}/\theta)^2$. For lower-temperature models, $k^{*}$ ranged from 5 to 18 runs; the two criteria agree on the ordering of models and temperatures, and, critically, the partial $r$ at $k = K$ and at $k = k^{*}$ did not differ significantly. For GPT-OSS-20B and Qwen3-32B at temp = 2.0, the token variability is so large that $k^{*}$ far exceeds our pool of available runs; for GPT-OSS-20B this corroborates our original finding of non-convergence, while for Qwen3-32B it reflects a stricter criterion than the bootstrap method, which found convergence at 24 runs.

\begin{table}[h]
\centering
\footnotesize
\setlength{\tabcolsep}{2pt}
\resizebox{\columnwidth}{!}{%
\begin{tabular}{lcccccc}
\toprule
\textbf{Model} & \textbf{Temp} & \textbf{$K$} & \textbf{$\bar{\sigma}$} & \textbf{$k^{*}$} & \textbf{Partial $r$ @ $k^{*}$} & \textbf{Partial $r$ @ $K$} \\
\midrule
R1  & 0.6 & 10 & 0.20 & 5          & +0.390 & +0.378 \\
R1  & 2.0 & 10 & 0.22 & 5          & +0.353 & +0.359 \\
GPT  & 1.0 & 25 & 0.42 & 18         & +0.561 & +0.554 \\
GPT  & 2.0 & 25 & 1.63 & $>268^\dagger$ & ---    & +0.238 \\
Qwen3    & 0.6 & 25 & 0.28 & 9          & +0.287 & +0.320 \\
Qwen3    & 2.0 & 25 & 0.76 & $>59^\dagger$  & ---    & +0.292 \\
\bottomrule
\end{tabular}}
\caption{\small $\bar{\sigma}$: median per-item SD of log(thinking tokens); $k^{*}$ computed from unrounded $\bar{\sigma}$ using the SE threshold above. $\dagger$ Pool exhausted}
\label{tab:cot-convergence}
\end{table}

\section{Stochastic Attenuation}
\label{appendix:attenuation}
\begin{figure}[H]
  \centering
    \includegraphics[width=\columnwidth]{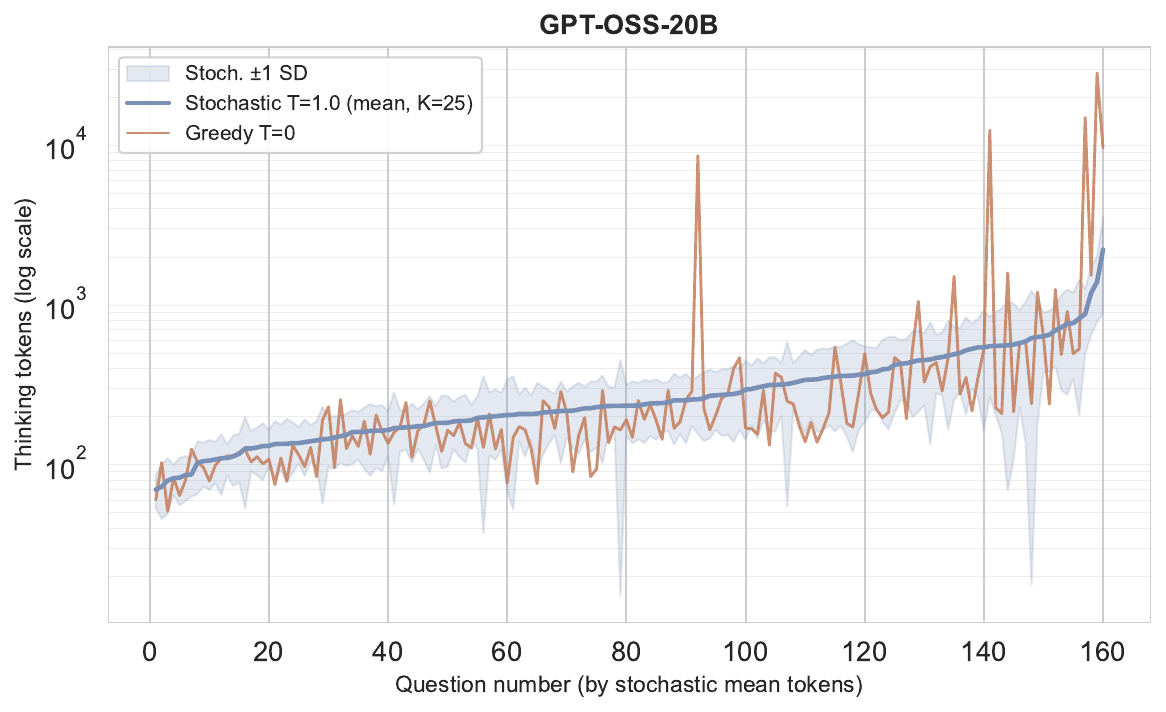}
  \caption{GPT-OSS-20B thinking tokens per item (log scale), with items sorted by stochastic mean token count (low to high difficulty).}
  \label{fig:stochastic-attenuation}
\end{figure}

\end{document}